\documentclass[11pt,a4paper]{article}
\usepackage[hyperref]{emnlp2020}
\usepackage{times}
\usepackage{url}
\usepackage{latexsym}
\usepackage{graphicx}
\usepackage{subfigure}
\usepackage{amsmath,amssymb}
\usepackage{booktabs} 
\usepackage{appendix}

\usepackage{microtype}

\aclfinalcopy 

\title{Multi-task Learning for Multilingual Neural Machine Translation}

\author{
    Yiren Wang$^{\dagger, }$\thanks{\ \ Work done while interning at Microsoft.} \ , 
    ChengXiang Zhai$^\dagger$, 
    Hany Hassan Awadalla$^\ddagger$ \\
    $^\dagger$University of Illinois at Urbana-Champaign \\ 
    $^\ddagger$Microsoft \\
    $^\dagger$\texttt{\{yiren, czhai\}@illinois.edu} \\
    $^\ddagger$\texttt{hanyh@microsoft.com}
}

\date{}

\begin{document}
\maketitle
\begin{abstract}
While monolingual data has been shown to be useful in improving bilingual neural machine translation (NMT), effectively and efficiently leveraging monolingual data for Multilingual NMT (MNMT) systems is a less explored area. In this work, we propose a multi-task learning (MTL) framework that jointly trains the model with the translation task on bitext data and two denoising tasks on the monolingual data. We conduct extensive empirical studies on MNMT systems with $10$ language pairs from WMT datasets. We show that the proposed approach can effectively improve the translation quality for both high-resource and low-resource languages with large margin, achieving significantly better results than the individual bilingual models. We also demonstrate the efficacy of the proposed approach in the zero-shot setup for language pairs without bitext training data. Furthermore, we show the effectiveness of MTL over pre-training approaches for both NMT and cross-lingual transfer learning NLU tasks; the proposed approach outperforms massive scale models trained on single task.

\end{abstract}

\section{Introduction}
Multilingual Neural Machine Translation (MNMT), which leverages a single NMT model to handle the translation of multiple languages, has drawn research attention in recent years~\cite{dong2015multi,firat2016multi,ha2016toward,johnson2017google,arivazhagan2019massively}. MNMT is appealing since it greatly reduces the cost of training and serving separate models for different language pairs~\cite{johnson2017google}. It has shown great potential in knowledge transfer among languages, improving the translation quality for low-resource and zero-shot language pairs~\cite{zoph2016transfer,firat2016zero,arivazhagan2019massively}.

Previous works on MNMT has mostly focused on model architecture design with different strategies of parameter sharing~\cite{firat2016multi,blackwood2018multilingual,sen2019multilingual} or representation sharing~\cite{gu2018universal}. Existing MNMT systems mainly rely on bitext training data, which is limited and costly to collect. Therefore, effective utilization of monolingual data for different languages is an important research question yet is less studied for MNMT.

Utilizing monolingual data (more generally, the unlabeled data) has been widely explored in various NMT and natural language processing (NLP) applications. Back translation (BT)~\cite{sennrich2016improving}, which leverages a target-to-source model to translate the target-side monolingual data into source language and generate pseudo bitext, has been one of the most effective approaches in NMT. However, well trained NMT models are required to generate back translations for each language pair, it is computationally expensive to scale in the multilingual setup. Moreover, it is less applicable to low-resource language pairs without adequate bitext data. Self-supervised pre-training approaches~\cite{radford2018improving,devlin2019bert,conneau2019cross,lewis2019bart,liu2020multilingual}, which train the model with denoising learning objectives on the large-scale monolingual data, have achieved remarkable performances in many NLP applications. However, catastrophic forgetting effect~\cite{thompson2019overcoming}, where finetuning on a task leads to degradation on the main task, limits the success of continuing training NMT on models pre-trained with monolingual data. Furthermore, the separated pre-training and finetuning stages make the framework less flexible to introducing additional monolingual data or new languages into the MNMT system.

In this paper, we propose a multi-task learning (MTL) framework to effectively utilize monolingual data for MNMT. Specifically, the model is jointly trained with translation task on multilingual parallel data and two auxiliary tasks: masked language modeling (MLM) and denoising auto-encoding (DAE) on the source-side and target-side monolingual data respectively. We further present two simple yet effective scheduling strategies for the multilingual and multi-task framework. In particular, we introduce a dynamic temperature-based sampling strategy for the multilingual data. To encourage the model to keep learning from the large-scale monolingual data, we adopt dynamic noising ratio for the denoising objectives to gradually increase the difficulty level of the tasks. 

We evaluate the proposed approach on a large-scale multilingual setup with $10$ language pairs from the WMT datasets. We study three English-centric multilingual systems, including many-to-English, English-to-many, and many-to-many. We show that the proposed MTL approach significantly boosts the translation quality for both high-resource and low-resource languages. Furthermore, we demonstrate that MTL can effectively improve the translation quality on zero-shot language pairs with no bitext training data. In particular, MTL achieves even better performance than the pivoting approach for multiple low-resource language pairs. We further show that MTL outperforms pre-training approaches on both NMT tasks as well as cross-lingual transfer learning for NLU tasks, despite being trained on very small amount of data in comparison to pre-training approaches.

The contributions of this paper are as follows. First, we propose a new MTL approach to effectively utilize monolingual data for MNMT. Second, we introduce two simple yet effective scheduling strategies, namely the dynamic temperature-based sampling and dynamic noising ratio strategy. Third, we present detailed ablation studies to analyze various aspects of the proposed approach. Finally, we demonstrate for the first time that MNMT with MTL models can be effectively used for cross-lingual transfer learning for NLU tasks with similar or better performance than the state-of-the-art massive scale pre-trained models using single task. 

\section{Background}

\paragraph{Neural Machine Translation} NMT adopts the sequence-to-sequence framework, which consists of an encoder and a decoder network built upon deep neural networks~\cite{sutskever2014sequence,bahdanau2014neural,gehring2017convolutional,vaswani2017attention}. The input source sentence is mapped into context representations in a continuous representation space by the encoder, which are then fed into the decoder to generate the output sentence. Given a language pair $(x, y)$, the objective of the NMT model training is to maximize the conditional probability $P(y|x; \theta)$ of the target sentence given the source sentence. 

NMT heavily relies on high-quality and large-scale bitext data. Various strategies have been proposed to augment the limited bitext by leveraging the monolingual data. Back translation~\cite{sennrich2016improving} utilizes the target-side monolingual data. Self learning~\cite{zhang2016exploiting} leverages the source-side monolingual data. Dual learning paradigms utilize monolingual data in both source and target language~\cite{he2016dual,wang2018multiagent,wu2019exploiting}. While these approaches can effectively improve the NMT performance, they have two limitations. First, they introduce additional cost in model training and translation generation, and therefore are less efficient when scaling to the multilingual setting. Second, back translation requires a good baseline model with adequate bitext data to start from, which limits its efficiency on low-resource settings.

\paragraph{Multilingual NMT} MNMT aims to train a single translation model that translates between multiple language pairs~\cite{firat2016multi,johnson2017google}. Previous works explored the model architecture design with different parameter sharing strategies, such as partial sharing with shared encoder~\cite{dong2015multi,sen2019multilingual}, shared attention~\cite{firat2016multi}, task-specific attention~\cite{blackwood2018multilingual}, and full model sharing with language identifier~\cite{johnson2017google,ha2016toward,arivazhagan2019massively}. There are also extensive studies on representation sharing that shares lexical, syntactic, or sentence level representations across different languages~\cite{zoph2016transfer,nguyen2017transfer,gu2018universal}. The models in these works rely on bitext for training, and the largely available monolingual data has not been effectively leveraged.

\paragraph{Self-supervised Learning} This work is motivated by the recent success of self-supervised learning for NLP applications~\cite{radford2018improving,devlin2019bert,lample2018unsupervised,lample2018phrase,conneau2019cross,lewis2019bart,liu2020multilingual}. Different denoising objectives have been designed to train the neural networks on large-scale unlabeled text. In contrast to previous work in pre-training with separated self-supervised pre-training and supervised finetuning stages, we focus on a multi-task setting to \emph{jointly} train the MNMT model on both bitext and monolingual data.

\paragraph{Multi-task Learning} Multi-task learning (MTL) \cite{caruana1997multitask}, which trains the model on several related tasks to encourage representation sharing and improve generalization performance, has been successfully used in many different machine learning applications~\cite{collobert2008unified,deng2013new,ruder2017overview}. In the context of NMT, MTL has been explored mainly to inject linguistic knowledge~\cite{luong2015multi,niehues2017exploiting,eriguchi2017learning,zaremoodi2018neural,kiperwasser2018scheduled} with tasks such as part-of-speech tagging, dependency parsing, semantic parsing, etc. In this work, we instead focus on auxiliary self-supervised learning tasks to leverage the monolingual data.

\section{Approach}

\begin{figure*}[t]
\label{fig:obj}
\centering
\begin{minipage}{0.45\linewidth}
\subfigure[Masked Language Model (MLM)]{
    \centering
	\label{subfig:obj-mlm}
	\includegraphics[width=0.85\linewidth]{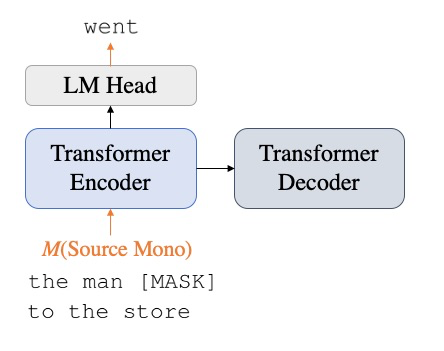} } 
\end{minipage}%
\begin{minipage}{0.45\linewidth}
\subfigure[Denoising Auto-Encoding (DAE)]{
    \centering
	\label{subfig:obj-dae}
	\includegraphics[width=0.85\linewidth]{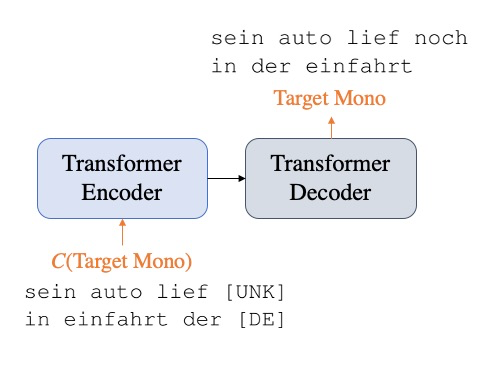} }
\end{minipage}
\caption{Illustration of the auxiliary tasks with monolingual data}
\end{figure*}

\subsection{Multi-task Learning}
\label{subsec:mtl_obj}

The main task in the MTL framework is the translation task trained on bitext corpora $D_B$ of sentence pairs $(x, y)$ with the cross-entropy loss:
\begin{equation}
\label{eq_ce}
\mathcal{L}_{MT}=\mathbb{E}_{(x,y) \sim D_B}[- \log P(y|x)] 
\end{equation}

With the large amount of monolingual data in different languages, we can train language models on both source-side~\footnote{For the English-to-Many translation model, the source-side language is English; for Many-to-English and Many-to-Many, it refers the set of all other languages. Similarly for the target-side language.} and target-side languages. We introduce two denoising language modeling tasks to help improve the quality of the translation model: the masked language model (MLM) task and the denoising auto-encoding (DAE) task.

\paragraph{Masked Language Model} In the masked language model (MLM) task~\cite{devlin2019bert}, sentences with tokens randomly masked are fed into the model and the model attempts to predict the masked tokens based on their context. MLM is beneficial for learning deep bidirectional representations. We introduce MLM as an auxiliary task to improve the quality of the encoder representations especially for the low-resource languages. 
As is illustrated in Figure~\ref{subfig:obj-mlm}, we add an additional output layer to the encoder of the translation model and train the encoder with MLM on source-side monolingual data. The output layer is dropped during inference. The cross entropy loss for predicting the masked tokens is denoted as $\mathcal{L}_{MLM}$. 

Following BERT~\cite{devlin2019bert}, we randomly sample $R_{M}\%$ units in the input sentences and replace them with a special \texttt{[MASK]} token. A unit can either be a subword token, or a word consists of one or multiple subword tokens. We refer to them as token-level and word-level MLM. 

\paragraph{Denoising Auto-Encoding (DAE)} Denoising auto-encoding (DAE)~\cite{vincent2008extracting} has been demonstrated to be an effective strategy for unsupervised NMT~\cite{lample2018unsupervised,lample2018phrase}. Given a monolingual corpus $D_M$ and a stochastic noising model $C$, DAE minimizes the reconstruction loss as shown in Eqn~\ref{eq_dae}:
\begin{equation}
\label{eq_dae}
\mathcal{L}_{DAE}=\mathbb{E}_{x \sim D_M}[- \log P(x|C(x))] 
\end{equation}

As is illustrated in Figure~\ref{subfig:obj-dae}, we train all model parameters with DAE on the target-side monolingual data. Specifically, we feed the target-side sentence to the noising model $C$ and append the corresponding language ID symbol; the model then attempts to reconstruct the original sentence. 

We introduce three types of noises for the noising model $C$. 1) \textit{Text Infilling}~\cite{lewis2019bart}: Following~\cite{liu2020multilingual}, we randomly sample $R_{D}\%$ text spans with span lengths drawn from a Poisson distribution ($\lambda=3.5$). We replace all words in each span with a single blanking token. 2) \textit{Word Drop \& Word Blank}: we randomly sample words from each input sentence, which are either removed or replaced with blanking tokens for each token position. 3) \textit{Word Swapping}: we slightly shuffle the order of words in the input sentence. Following~\cite{lample2018unsupervised}, we apply a random permutation $\sigma$ with condition $|\sigma(i)-i| \leq k, \forall i \in \{1,n\}$, where $n$ is the length of the input sentence, and $k=3$ is the maximum swapping distance.

\paragraph{Joint Training} In the training process, the two self-learning objectives are combined with the cross-entropy loss for the translation task:
\begin{equation}
\mathcal{L}=\mathcal{L}_{MT} + \mathcal{L}_{MLM} + \mathcal{L}_{DAE}
\end{equation}
In particular, we use bitext data for the translation objective, source-side monolingual data for MLM, and target-side monolingual data for the DAE objective. A language ID symbol \texttt{[LID]} of the target language is appended to the input sentence in the translation and DAE tasks. 

\subsection{Task Scheduling}
\label{subsec:mtl_schedule}
The scheduling of tasks and data associated with the task is important for multi-task learning. We further introduce two simple yet effective scheduling strategies in the MTL framework.

\paragraph{Dynamic Data Sampling}
One serious yet common problem for MNMT is data imbalance across different languages. Training the model with the true data distribution would starve the low-resource language pairs. Temperature-based batch balancing~\cite{arivazhagan2019massively} is demonstrated to be an effective heuristic to ease the problem. For language pair $l$ with bitext corpus $D_l$, we sample instances with probability proportional to $(\frac{|D_l|}{\sum_k |D_k|})^{\frac{1}{T}}$, where $T$ is the sampling temperature. 

While MNMT greatly improves translation quality for low-resource languages, performance deterioration is generally observed for high resource languages. One hypothesized reason is that the model might converge before well trained on high-resource data~\cite{bapna2019simple}. To alleviate this problem, we introduce a simple heuristic to feed more high-resource language pairs in the early stage of training and gradually shift more attention to the low-resource languages. To achieve this, we modify the sampling strategy by introducing  dynamic sampling temperature $T(k)$ as a function of the number of training epochs $k$. We use a simple linear functional form for $T(k)$:
\begin{equation}
T(k) = \text{min}\left(T_m, (k-1)\frac{T_m-T_0}{N}+T_0\right)
\label{eqn:temp}
\end{equation}
Where $T_0$ and $T_m$ are  the initial and maximum value for sampling temperature respectively. $N$ is the number of warm-up epochs. The sampling temperature starts from a smaller value $T_0$, resulting in sampling leaning towards true data distribution. $T(k)$ gradually increases in the training process to encourage over-sampling low-resource languages more to avoid them getting starved. 

\paragraph{Dynamic Noising Ratio} We further schedule the difficulty level of MLM and DAE from easier to more difficult. The main motivation is that training algorithms perform better when starting with easier tasks and gradually move to harder ones as promoted in curriculum learning~\cite{elman1993learning}. Furthermore, increasing the learning difficulty can potentially help  avoid saturation and encourage the model to keep learning from abundant data. 

Given the monolingual data, the difficulty level of MLM and DAE tasks mainly depends on the noising ratio. Therefore, we introduce dynamic noising ratio $R(k)$ as a function of training steps:
\begin{equation}
R(k) = \text{min}\left(R_m, (k-1)\frac{R_m-R_0}{M}+R_0\right)
\label{eqn:noise}
\end{equation}
Where $R_0$ and $R_m$ are the lower and upper bound for noising ratio respectively and $M$ is the number of warm-up epochs. Noising ratio $R$ refers to the masking ratio $R_M$ in MLM and the blanking ratio $R_D$ of the \textit{Text Infilling} task for DAE.

\section{Experimental Setup}

\subsection{Data}
\label{subsec:data}

We evaluate MTL on a multilingual setting with 10 languages to and from English (En), including French (Fr), Czech (Cs), German (De), Finnish (Fi), Latvian (Lv), Estonian (Et), Romanian (Ro), Hindi (Hi), Turkish (Tr) and Gujarati (Gu). 

\paragraph{Bitext Data} The bitext training data comes from the WMT corpus. Detailed desciption and statistics can be found in Appendix~\ref{appen:bitext}.

\paragraph{Monolingual Data} The monolingual data we use is mainly from NewsCrawl\footnote{\url{http://data.statmt.org/news-crawl/}}. We apply a series of filtration rules to remove the low-quality sentences, including duplicated sentences, sentences with too many punctuation marks or invalid characters, sentences with too many or too few words, etc. We randomly select $5$M filtered sentences for each language. For low-resource languages without enough sentences from NewsCrawl, we leverage data from CCNet~\cite{wenzek2019ccnet}.

\paragraph{Back Translation} We use the target-to-source bilingual models to back translate the target-side monolingual sentences into the source domain for each language pair. The synthetic parallel data from back translation is mixed and shuffled with bitext and used together for the translation objective in training. We use the same monolingual data for back translation as the multi-task learning in all our experiments for fair comparison.

\subsection{Model Configuration} 
\label{subsec:exp_hparams}
We use Transformer for all our experiments using the PyTorch implementation\footnote{\url{https://github.com/pytorch/fairseq}}~\cite{ott2019fairseq}. We adopt the \texttt{transformer\_big} setting~\cite{vaswani2017attention} with a $6$-layer encoder and decoder. The dimensions of word embeddings, hidden states, and non-linear layer are set as $1024$, $1024$ and $4096$ respectively, the number of heads for multi-head attention is set as $16$. We use a smaller model setting for the bilingual models on low-resource languages Tr, Hi and Gu (with $3$ encoder and decoder layers, $256$ embedding and hidden dimension) to avoid overfitting and acquire better performance. 

We study three multilingual translation scenarios including many-to-English (X$\to$En), English-to-many (En$\to$X) and many-to-many (X$\to$X). For the multilingual model, we adopt the same Transformer architecture as the bilingual setting, with parameters fully shared across different language pairs. A target language ID token is appended to each input sentence. 


\subsection{Training and Evaluation}

All models are optimized with Adam~\cite{kingma2014adam} with $\beta_1=0.9$, $\beta_2=0.98$. We set the learning rate schedule following~\cite{vaswani2017attention} with initial learning rate $5\times10^{-4}$. Label smoothing~\cite{szegedy2016rethinking} is adopted with $0.1$. The models are trained on $8$ V100 GPUs with a batch size of $4096$ and the parameters are updated every $16$ batches. 
During inference, we use beam search with a beam size of $5$ and length penalty $1.0$. The BLEU score is measured by the de-tokenized case-sensitive SacreBLEU\footnote{SacreBLEU signatures: BLEU+case.mixed+lang.\$l1-\$l2\\numrefs.1+smooth.exp+test.\$SET+tok.13a+version.1.4.3, where \$l1, \$l2 are the language code (Table~\ref{tbl:data-stats}), \$SET is the corresponding test set for the language pair.}~\cite{post2018call}. 

\begin{table*}[t]
\centering
\small
\begin{tabular}{lcccccccccc}
\toprule
            & Fr      & Cs      & De      & Fi      & Lv      & Et      & Ro      & Hi      & Tr      & Gu      \\
Test Set    & wmt$15$ & wmt$18$ & wmt$18$ & wmt$18$ & wmt$17$ & wmt$18$ & wmt$16$ & wmt$14$ & wmt$18$ & wmt$19$ \\ 
\midrule
Bilingual   & $36.2$  & $28.5$  & $40.2$  & $19.2$  & $17.5$  & $19.7$  & $29.8$  & $14.1$  & $15.1$  & $9.3$   \\ 
\midrule
X $\to$ En  & $34.6$  & $28.0$  & $39.7$  & $20.1$  & $19.6$  & $23.9$  & $33.2$  & $20.5$  & $21.3$  & $16.1$  \\
+ MTL       & $36.4$  & $31.5$  & $42.3$  & $23.0$  & $22.1$  & $28.7$  & $37.9$  & $24.8$  & $25.7$  & $22.3$  \\
+ BT        & $35.3$  & $31.2$  & $44.3$  & $23.4$  & $21.4$  & $29.2$  & $37.9$  & $27.2$  & $25.5$  & $21.5$  \\
+ BT + MTL  & $35.3$  & $31.9$  & $45.4$  & $23.8$  & $22.4$  & $30.5$  & $39.1$  & $28.7$  & $27.6$  & $23.5$  \\ 
\midrule
X $\to$ X   & $33.9$  & $28.1$  & $39.0$  & $19.9$  & $19.5$  & $24.5$  & $33.7$  & $22.4$  & $22.0$  & $17.2$  \\
+ MTL       & $35.1$  & $29.6$  & $40.1$  & $21.7$  & $21.3$  & $27.3$  & $36.8$  & $23.9$  & $25.2$  & $23.3$  \\ 
+ BT        & $34.3$  & $30.6$  & $43.7$  & $22.8$  & $20.9$  & $28.0$  & $37.3$  & $26.4$  & $25.5$  & $22.5$  \\
+ BT + MTL  & $35.3$  & $31.2$  & $43.7$  & $23.1$  & $21.5$  & $29.5$  & $38.1$  & $27.5$  & $26.2$  & $23.4$  \\
\bottomrule
\end{tabular}
\caption{BLEU scores of $10$ languages $\to$ English translation with bilingual, X$\to$En and X$\to$X systems. The languages are arranged from high-resource (left) to low-resource (right). \label{tbl:results_x_en}}
\end{table*}

\begin{table*}[t]
\centering
\small
\begin{tabular}{lcccccccccc}
\toprule
            & Fr      & Cs      & De      & Fi      & Lv      & Et      & Ro      & Hi      & Tr      & Gu      \\
Test Set    & wmt$15$ & wmt$18$ & wmt$18$ & wmt$18$ & wmt$17$ & wmt$18$ & wmt$16$ & wmt$14$ & wmt$18$ & wmt$19$ \\
\midrule
Bilingual   & $36.3$  & $22.3$  & $40.2$  & $15.2$  & $16.5$  & $15.0$  & $23.0$  & $12.2$  & $13.3$  & $7.9$   \\ 
\midrule
En $\to$ X  & $33.5$  & $20.8$  & $39.0$  & $14.9$  & $18.0$  & $19.8$  & $25.5$  & $12.4$  & $15.7$  & $11.9$  \\
+ MTL       & $33.8$  & $21.7$  & $39.8$  & $15.2$  & $18.5$  & $21.1$  & $26.5$  & $16.1$  & $17.6$  & $15.4$  \\
+ BT        & $35.9$  & $22.5$  & $41.5$  & $17.3$  & $21.8$  & $23.0$  & $28.8$  & $19.1$  & $18.6$  & $15.5$  \\
+ BT + MTL  & $36.1$  & $23.6$  & $42.0$  & $17.7$  & $22.4$  & $24.0$  & $29.8$  & $19.8$  & $19.4$  & $17.8$  \\ 
\midrule
X $\to$ X   & $32.2$  & $19.4$  & $37.3$  & $14.5$  & $17.5$  & $19.6$  & $25.4$  & $13.9$  & $16.3$  & $12.0$  \\
+ MTL       & $33.3$  & $20.9$  & $39.2$  & $15.6$  & $19.3$  & $21.1$  & $26.8$  & $16.5$  & $18.1$  & $15.5$  \\ 
+ BT        & $35.9$  & $22.0$  & $40.0$  & $16.3$  & $21.1$  & $22.8$  & $28.7$  & $19.0$  & $18.2$  & $15.9$  \\
+ BT + MTL  & $35.8$  & $22.4$  & $41.2$  & $16.9$  & $21.7$  & $23.2$  & $29.7$  & $19.2$  & $18.7$  & $16.0$  \\
\bottomrule
\end{tabular}
\caption{BLEU scores of English $\to 10$ languages translation with bilingual, En$\to$X and X$\to$X systems. The languages are arranged from high-resource (left) to low-resource (right). \label{tbl:results_en_x}}
\end{table*}

\section{Results}

\subsection{Main Results}
\label{subsec:main_results}

We compare the performance of the bilingual models (\textit{Bilingual}), multilingual models trained on bitext only, trained on both bitext and back translation (\textit{+BT}) and trained with the proposed multi-task learning (\textit{+MTL}). Translation results of the $10$ languages translated to and from English are presented in Table~\ref{tbl:results_x_en} and \ref{tbl:results_en_x} respectively. We can see that:

\noindent 1. \textit{Bilingual vs. Multilingual}: The multilingual baselines perform better on lower-resource languages, but perform  worse than individual bilingual models on high-resource languages like Fr, Cs and De. This is in concordance with the previous observations~\cite{arivazhagan2019massively} and is consistent across the three multilingual systems (i.e., X$\to$En, En$\to$X and X$\to$X).

\noindent 2. \textit{Multi-task learning}: Models trained with multi-task learning (+MTL) significantly outperform the multilingual baselines for all the languages pairs in all three multilingual systems, demonstrating the effectiveness of the proposed framework.

\noindent 3. \textit{Back Translation}: With the same monolingual corpus, MTL achieves better performance on some language pairs (e.g. Fr$\to$En, Gu$\to$En), while getting outperformed on some others, especially on the En$\to$X direction. However, back translation is computationally expensive as it involves the additional procedure of training $10$ bilingual models ($20$ for the X$\to$X system) and generating translations for each monolingual sentence. Combining MTL with BT (+BT+MTL) introduces further improvements for most language pairs without using any additional monolingual data. This suggests that when there is enough computation budget for BT, MTL can still be leveraged  to provide good complementary improvement.

\subsection{Zero-shot Translation}
\label{subsec:zero_shot}
We further evaluate the proposed approach on zero-shot translation of non English-centric language pairs. We compare the performances of the pivoting method, the X$\to$X baseline system, X$\to$X with BT, and with MTL. For the pivoting method, the source language is translated into English first, and then translated into the target language~\cite{de2006catalan,utiyama2007comparison}. We evaluate on a group of high-resource languages with a multi-way parallel test set for De, Cs, Fr and En, constructed by newstest2009 with $3027$ sentences and that of a group of low-resource languages Et, Hi, Tr and Hi ($995$ sentences). The results are shown in Table~\ref{tbl:result_zero_high} and~\ref{tbl:result_zero_low} respectively. 

\begin{table}[]
\centering
\small
\begin{tabular}{lcccc}
\toprule
            & De$\to$Fr & Fr$\to$De & Cs$\to$De & De$\to$Cs \\ 
\midrule
Pivoting    & $22.1$    & $19.1$    & $17.5$    & $15.9$    \\ 
\midrule
X$\to$X     & $15.1$    & $11.9$    & $15.5$    & $15.2$    \\
+ BT        & $19.7$    & $7.4$     & $17.0$    & $7.8$     \\
{\footnotesize + BT + MTL} & $20.1$    & $12.2$    & $19.7$    & $12.0$    \\ 
\bottomrule
\end{tabular}
\caption{Zero-shot translation performances on high-resource language pairs.}
\label{tbl:result_zero_high}
\end{table}

Utilizing monolingual data with MTL significantly improves the zero-shot translation quality of the X$\to$X system, further demonstrating the effectiveness of the proposed approach. In particular, MTL achieves significantly better results than the pivoting approach on the high-resource pair Cs$\to$De and almost all low-resource pairs. Furthermore, leveraging monolingual data through BT does not perform well for many low-resource language pairs, resulting in comparable and even downgraded performances. We conjecture that this is related to the quality of the back translations. MTL helps overcome such limitations with the auxiliary self-supervised learning tasks.

\subsection{MTL vs. Pre-training}

\begin{table}[]
\centering
\small
\begin{tabular}{lcccc}
\toprule
            & Et$\to$Hi & Hi$\to$Et & Hi$\to$Tr & Tr$\to$Hi        \\
\midrule
Pivoting    & $8.1$     & $7.1$     & $3.9$     & $6.0$            \\
X$\to$X     & $5.0$     & $5.4$     & $2.5$     & $3.6$            \\
+ BT        & $4.9$     & $5.5$     & $4.5$     & $5.7$            \\
{\footnotesize + BT + MTL}      & $9.3$     & $8.1$     & $5.7$     & $9.4$            \\
\midrule \midrule
            & Et$\to$Tr & Tr$\to$Et & Hi$\to$Lv & Lv$\to$Hi        \\
\midrule
Pivoting    & $7.1$     & $7.8$     & $8.6$     & $7.9$            \\
X$\to$X     & $7.8$     & $8.8$     & $6.1$     & $4.7$            \\
+ BT        & $7.3$     & $7.5$     & $7.3$     & $6.1$            \\
{\footnotesize + BT + MTL}      & $7.8$     & $10.5$    & $8.2$     & $8.6$            \\
\bottomrule
\end{tabular}
\caption{Zero-shot translation performances on low-resource language pairs.}
\label{tbl:result_zero_low}
\end{table}

\begin{figure}[t]
    \centering
    \includegraphics[width=0.95\linewidth]{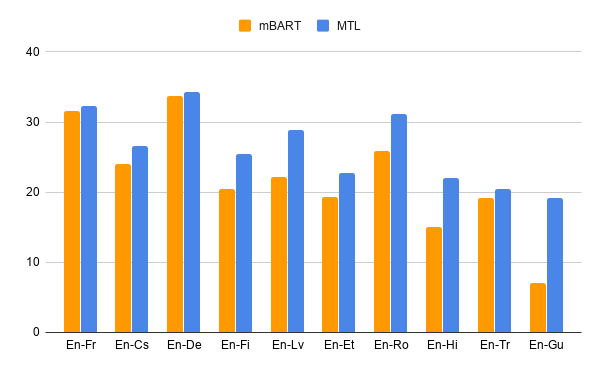}
    \caption{Comparison with mBART on En$\to$X language pairs. BLEU scores are reported on the full individual validation set.}
    \label{fig:vs_mbart}
\end{figure}

We also compare MTL with mBART~\cite{liu2020multilingual}, the state-of-the-art multilingual pre-training method for NMT. We adopt the officially released mBART model pre-trained on CC25 corpus\footnote{\url{https://dl.fbaipublicfiles.com/fairseq/models/mbart/mbart.CC25.tar.gz}} and finetune the model on the same bitext training data used in MTL for each language pair. As shown in Figure~\ref{fig:vs_mbart}, MTL outperforms mBART on all language pairs. This suggests that in the scenario of NMT, jointly training the model with MT task and self-supervised learning tasks could be a better task design than the separated pre-training and finetuning stages. It is worth noting that mBart is utilizing much more monolingual data; for example, it uses   $55$B  English tokens and $10$B French tokens, while our approach is using just $100$M tokens each. This indicates that MTL is more data efficient.

\begin{table}[t]
\small
\centering
\begin{tabular}{lcccc}
\toprule
             & \multicolumn{2}{c}{De$\to$En} & \multicolumn{2}{c}{Tr$\to$En} \\
Systems      & X$\to$En       & X$\to$X      & X$\to$En       & X$\to$X      \\
\midrule
Multilingual & $36.5$         & $36.1$       & $20.0$         & $20.9$       \\
+ MLM        & $36.3$         & $36.6$       & $21.0$         & $21.4$       \\
+ DAE        & $37.7$         & $37.8$       & $21.7$         & $22.6$       \\
+ MLM + DAE  & $38.7$         & $37.6$       & $22.9$         & $23.7$       \\
\midrule
\midrule
             & \multicolumn{2}{c}{En$\to$De} & \multicolumn{2}{c}{En$\to$Tr} \\
Systems      & En$\to$X       & X$\to$X      & En$\to$X       & X$\to$X      \\
\midrule
Multilingual & $33.0$         & $32.0$       & $16.4$         & $17.0$       \\
+ MLM        & $32.9$         & $32.6$       & $16.9$         & $17.2$       \\
+ DAE        & $33.7$         & $33.7$       & $17.3$         & $18.2$       \\
+ MLM + DAE  & $34.2$         & $33.6$       & $18.0$         & $18.3$       \\
\bottomrule
\end{tabular}
\caption{Comparison of different multi-task learning objectives on De-En and Tr-En translation. BLEU scores are reported on the full individual validation set. \label{tbl:study_obj}}
\end{table}

\subsection{Multi-task Objectives}
\label{subsec:study_mtl_obj}
We present ablation study on the learning objectives of the multi-task learning framework. We compare performance of multilingual baseline model with translation objective only, jointly learning translation with MLM, jointly learning translation with DAE, and the combination of all objectives. 
Table~\ref{tbl:study_obj} shows the results on a high-resource pair De$\leftrightarrow$En and low-resource pair Tr$\leftrightarrow$En. We can see that introducing MLM or DAE can both effectively improve the performance of multilingual systems, and the combination of both yields the best performance. We also observe that MLM is more beneficial for `$\to$En' compared with `En$\to$' direction, especially for the low-resource languages. This is in concordance with our intuition that the MLM objective contributes to improving the encoder quality and source-side language modeling for low-resource languages. 

\begin{figure}[t]
    \centering
    \includegraphics[width=0.95\linewidth]{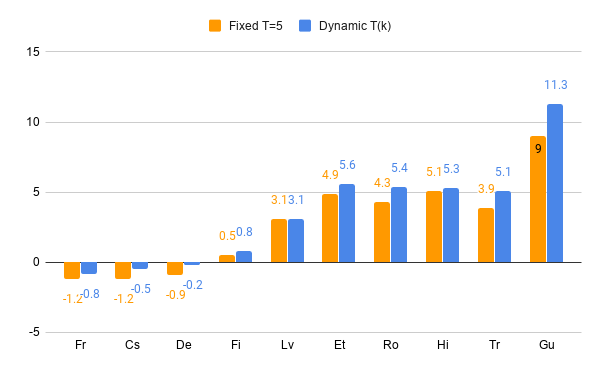}
    \caption{Performance gain of data sampling strategies on the X$\to$En system. Results are reported as $\Delta$BLEU relative to the corresponding bilingual baseline on validation sets. The languages are arranged from high-resource (left) to low-resource (right).}
    \label{fig:temp_x_en}
\end{figure}

\begin{figure}[t]
    \centering
    \includegraphics[width=0.95\linewidth]{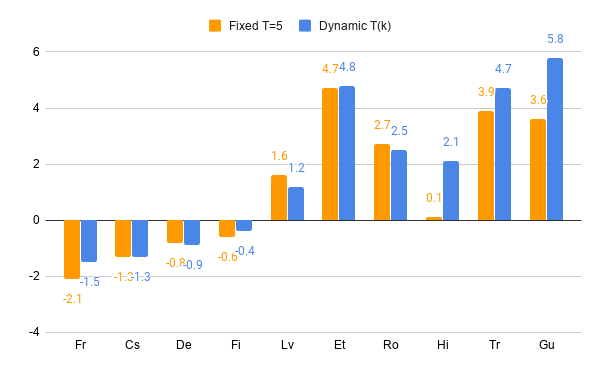}
    \caption{Performance of data sampling strategies on the En$\to$X system.}
    \label{fig:temp_en_x}
\end{figure}

\begin{figure}[t]
    \centering
    \includegraphics[width=0.95\linewidth]{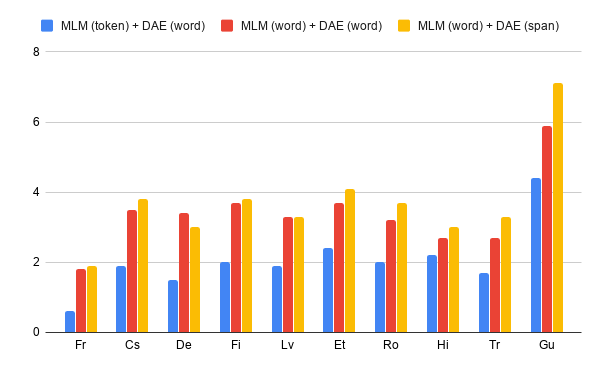}
    \caption{Performance of different noising schemes on the X$\to$En system. Results are reported as $\Delta$BLEU relative to the multilingual X$\to$En baseline on validation sets. The languages are arranged from high-resource (left) to low-resource (right).}
    \label{fig:noise_scheme}
\end{figure}

\subsection{Dynamic Sampling Temperature}
\label{subsec:study_temp}
We study the effectiveness of the proposed dynamic sampling strategy. We compare multilingual systems using a fixed sampling temperature $T=5$ with systems using dynamic temperature $T(k)$ defined in Equation~\ref{eqn:temp}. We set $T_0=1, T_m=5, N=5$, which corresponds to gradually increasing the temperature from $1$ to $5$ with $5$ training epochs  and saturate to $T=5$ afterwards. The results for X$\to$En and En$\to$X systems are presented in Figure~\ref{fig:temp_x_en} and~\ref{fig:temp_en_x} respectively, where we report $\Delta$BLEU relative to their corresponding bilingual baseline model that was evaluated on the individual validation sets for each language pairs. The dynamic temperature strategy improves the quality for high-resource language pairs (e.g. Fr$\to$En, De$\to$En, En$\to$Fr), while introducing minimum effect for mid-resource languages (Lv). Surprisingly, the proposed strategy also greatly boosts performance for low-resource languages Tr and Gu, with over $+1$ BLEU gain for both to and from English direction. 

\subsection{Noising Scheme}
\label{subsec:study_noise}
We study the effect of different noising schemes in the MLM and DAE objectives. As introduced in Section~\ref{subsec:mtl_obj}, we have token-level and word-level masking scheme for MLM depending on the unit of masking. We also have two noising schemes for DAE, where the \textit{Text Infilling} task blanks a span of words (span-level), and the \textit{Word Blank} task blanks the input sentences at word-level. We compare performance of these different noising schemes  on X$\to$En system as  shown in Figure~\ref{fig:noise_scheme}.

We report $\Delta$BLEU relative to the multilingual X$\to$En baseline on the corresponding language pairs for each noising scheme. As we can see, the model benefits most from the word-level MLM and the span-level \textit{Text Infilling} task for DAE. This is in concordance with the intuition that the \textit{Text Infilling} task teaches the model to predict the length of masked span and the exact tokens at the same time, making it a harder task to learn. We use the word-level MLM and span-level DAE as the best recipe for our MTL framework.

\subsection{Noising Ratio Scheduling}
\label{subsec:study_noise_ratio}

In our initial experiments, we found that the dynamic noising ratio strategy does not effectively improve the performance. We suspect that it is due to the limitation of data scale. We experiment with a larger scale setting by increasing the amount of monolingual data from $5$M sentences for each language to $20$M. For low-resource languages without enough data, we take the full available amount ($18$M for Lv, $11$M for Et, $5.2$M for Gu). 

Table~\ref{tbl:result_dyn_noise} shows results on X$\to$En MNMT model with large-scale monolingual data setting. We compare the performance of multilingual with back translation baseline, a model with MTL and a model with both  MTL and  dynamic noising ratio. For the dynamic noising ratio, we set the masking ratio for MLM to increase from $10\%$ to $20\%$ and blanking ratio for DAE to increase from $20\%$ to $40\%$. As we can see, the dynamic noising strategy helps boost performance for mid-resource languages like Lv and Et, while introducing no negative effect to other languages. For future study, we would like to cast the dynamic noising ratio over different subsets of monolingual datasets to prevent the model from learning to copy and memorize.

\begin{table}[]
\centering
\small
\begin{tabular}{lccccc}
\toprule
                  & De     & Lv     & Et     & Hi     & Tr     \\
\midrule
Bilingual         & $32.9$ & $23.4$ & $18.7$ & $12.9$ & $16.1$ \\
\midrule
X$\to$En + BT     & $35.2$ & $30.1$ & $28.3$ & $18.7$ & $24.7$ \\
+ MTL             & $36.9$ & $31.9$ & $31.4$ & $21.6$ & $27.4$ \\
+ Dynamic         & $37.0$ & $32.4$ & $32.0$ & $21.7$ & $27.5$ \\
\bottomrule
\end{tabular}
\caption{BLEU scores of dynamic noising strategy on X$\to$En translation system with large-scale monolingual data setting on validation sets. \label{tbl:result_dyn_noise}}
\end{table}

\subsection{MTL for Cross-Lingual Transfer Learning for NLU}
\label{sec:nlu_res}

Large scale pre-trained cross-lingual language models such as mBERT~\cite{devlin2019bert} and XLM-Roberta~\cite{ConneauKGCWGGOZ20} are the state-of-the-art for cross-lingual transfer learning on natural language understanding (NLU) tasks, such as XNLI~\cite{xnli} and XGLUE~\cite{XGLUE}. Such models are trained on massive amount of monolingual data from all language as a masked language model. It has been shown that massive MNMT models are not able to match the performance of pre-trained language models such as XLM-Roberta on NLU downstream tasks~\cite{Siddhant2020}. In~\citet{Siddhant2020}, the MNMT models are massive scale models trained only on the NMT task. They are not able to outperform XLM-Roberta, which is trained with MLM task without any parallel data. In this work, we evaluate the effectiveness of our proposed MTL approach for cross-lingual transfer leaning on NLU application. Intuitively, MTL can bridge this gap since it utilizes NMT, MLM and DAE objectives. 

In the experiment, we train a system on $6$ languages using both bitext and monolingual data. For the bitext training data, we use $30$M parallel sentences per language pair from in-house data crawled from the web. For the monolingual data, we use $40$M sentences per language from CCNet~\cite{wenzek2019ccnet}. Though this is a relatively large-scale setup, it only leverages a fraction of the data used to train XLM-Roberta for those languages. We train the model with $12$ layers encoders and $6$ layers decoder. The hidden dimension is $768$ and the number of heads is $8$. We tokenize all data with the SentencePiece model~\cite{kudo2018sentencepiece} with the vocabulary size of $64$K. We train a many-to-many MNMT system with three tasks described in Section~\ref{subsec:mtl_obj}: NMT, MLM, and DAE. Once the model is trained, we use the encoder only and discard the decoder. We add a feedforward layer for the downstream tasks.

\begin{table}[]
\centering
\small
\begin{tabular}{lcccc}
\toprule
              & EN     & ES     & DE     & FR      \\
\midrule
XLM-Roberta   & $84.7$ & $79.4$ & $77.4$ & $79.1$  \\
MMTE          & $79.6$ & $71.6$ & $68.2$ & $69.5$  \\
MTL           & $84.8$ & $80.1$ & $78.3$ & $79.8$  \\
\bottomrule
\end{tabular}
\caption{Evaluation on XNLI task, XLM-Roberta results is our reproduction of the results. Massively Multilingual
Translation Encoder (MMTE) is reported from~\cite{Siddhant2020}  \label{tbl:xnli_results} }
\end{table}

As shown in Table~\ref{tbl:xnli_results}, MTL outperform both XLM-Roberta and MMTE~\cite{Siddhant2020} which are trained on massive amount of data in comparison to our system. XLM-Roberta is trained only on MLM task and MMTE is trained only on NMT task. Our MTL system is trained on three tasks. The results clearly highlight the effectiveness of multi-task learning, and demonstrate that it can outperform single-task systems trained on massive amount of data. We observe the same pattern in Table~\ref{tbl:xglue_results} with XGLUE NER task, which outperforms SOTA XLM-Roberta model.

\begin{table}[]
\centering
\small
\begin{tabular}{lccc}
\toprule
                 & EN     & ES     & DE       \\
\midrule
XLM-Roberta      & $91.1$ & $76.5$ & $70.3$   \\
MTL              & $91.2$ & $77.0$ & $75.0$   \\

\bottomrule
\end{tabular}
\caption{Evaluation on XGLUE NER task, XLM-Roberta results is our reproduction of the results. \label{tbl:xglue_results} }
\end{table}

\section{Conclusion}
In this work, we propose a multi-task learning framework that jointly trains the model with the translation task on bitext data, the masked language modeling task on the source-side monolingual data and the denoising auto-encoding task on the target-side monolingual data. We explore data and noising scheduling approaches and demonstrate their efficacy for the proposed approach. We show that the proposed MTL approach can effectively improve the performance of MNMT on both high-resource and low-resource languages with large margin, and can also significantly improve the translation quality for zero-shot language pairs without bitext training data. We showed that the proposed approach is more effective than pre-training followed by finetuning for NMT. Furthermore, we showed the effectiveness of multitask learning for cross-lingual downstream tasks outperforming  SOTA larger models trained on single task.

For future work, we are interested in investigating the proposed approach in a scaled setting with more languages and a larger amount of monolingual data. Scheduling the different tasks and different types of data would be an interesting problem. Furthermore, we would also like to explore the most sample efficient strategy to add a new language to a trained MNMT system.

\section*{Acknowledgment} 
We would like to thank Alex Muzio for helping with zero-shot scoring  and useful discussions. We also would like to thank Dongdong Zhang for helping with mBART comparison and Felipe Cruz Salinas for helping with XNLI and XGLUE scoring.

\bibliographystyle{acl_natbib}
\bibliography{mybib}

\begin{thebibliography}{52}
\expandafter\ifx\csname natexlab\endcsname\relax\def\natexlab#1{#1}\fi

\bibitem[{Arivazhagan et~al.(2019)Arivazhagan, Bapna, Firat, Lepikhin, Johnson,
  Krikun, Chen, Cao, Foster, Cherry et~al.}]{arivazhagan2019massively}
Naveen Arivazhagan, Ankur Bapna, Orhan Firat, Dmitry Lepikhin, Melvin Johnson,
  Maxim Krikun, Mia~Xu Chen, Yuan Cao, George Foster, Colin Cherry, et~al.
  2019.
\newblock Massively multilingual neural machine translation in the wild:
  Findings and challenges.
\newblock \emph{arXiv preprint arXiv:1907.05019}.

\bibitem[{Bahdanau et~al.(2014)Bahdanau, Cho, and Bengio}]{bahdanau2014neural}
Dzmitry Bahdanau, Kyunghyun Cho, and Yoshua Bengio. 2014.
\newblock Neural machine translation by jointly learning to align and
  translate.
\newblock \emph{arXiv preprint arXiv:1409.0473}.

\bibitem[{Bapna and Firat(2019)}]{bapna2019simple}
Ankur Bapna and Orhan Firat. 2019.
\newblock Simple, scalable adaptation for neural machine translation.
\newblock In \emph{Proceedings of the 2019 Conference on Empirical Methods in
  Natural Language Processing}, pages 1538--1548.

\bibitem[{Blackwood et~al.(2018)Blackwood, Ballesteros, and
  Ward}]{blackwood2018multilingual}
Graeme Blackwood, Miguel Ballesteros, and Todd Ward. 2018.
\newblock Multilingual neural machine translation with task-specific attention.
\newblock In \emph{Proceedings of the 27th International Conference on
  Computational Linguistics}, pages 3112--3122.

\bibitem[{Caruana(1997)}]{caruana1997multitask}
Rich Caruana. 1997.
\newblock Multitask learning.
\newblock \emph{Machine learning}, pages 41--75.

\bibitem[{Collobert and Weston(2008)}]{collobert2008unified}
Ronan Collobert and Jason Weston. 2008.
\newblock A unified architecture for natural language processing: Deep neural
  networks with multitask learning.
\newblock In \emph{Proceedings of the 25th international conference on Machine
  learning}, pages 160--167.

\bibitem[{Conneau et~al.(2020)Conneau, Khandelwal, Goyal, Chaudhary, Wenzek,
  Guzm{\'{a}}n, Grave, Ott, Zettlemoyer, and Stoyanov}]{ConneauKGCWGGOZ20}
Alexis Conneau, Kartikay Khandelwal, Naman Goyal, Vishrav Chaudhary, Guillaume
  Wenzek, Francisco Guzm{\'{a}}n, Edouard Grave, Myle Ott, Luke Zettlemoyer,
  and Veselin Stoyanov. 2020.
\newblock \href {https://www.aclweb.org/anthology/2020.acl-main.747/}
  {Unsupervised cross-lingual representation learning at scale}.
\newblock In \emph{Proceedings of the 58th Annual Meeting of the Association
  for Computational Linguistics, {ACL} 2020, Online, July 5-10, 2020}, pages
  8440--8451. Association for Computational Linguistics.

\bibitem[{Conneau and Lample(2019)}]{conneau2019cross}
Alexis Conneau and Guillaume Lample. 2019.
\newblock Cross-lingual language model pretraining.
\newblock In \emph{Advances in Neural Information Processing Systems}, pages
  7059--7069.

\bibitem[{Conneau et~al.(2018)Conneau, Rinott, Lample, Williams, Bowman,
  Schwenk, and Stoyanov}]{xnli}
Alexis Conneau, Ruty Rinott, Guillaume Lample, Adina Williams, Samuel Bowman,
  Holger Schwenk, and Veselin Stoyanov. 2018.
\newblock \href {https://doi.org/10.18653/v1/D18-1269} {{XNLI}: Evaluating
  cross-lingual sentence representations}.
\newblock In \emph{Proceedings of the 2018 Conference on Empirical Methods in
  Natural Language Processing}, Brussels, Belgium. Association for
  Computational Linguistics.

\bibitem[{De~Gispert and Marino(2006)}]{de2006catalan}
Adri{\`a} De~Gispert and Jose~B Marino. 2006.
\newblock Catalan-english statistical machine translation without parallel
  corpus: bridging through spanish.
\newblock In \emph{Proc. of 5th International Conference on Language Resources
  and Evaluation (LREC)}, pages 65--68. Citeseer.

\bibitem[{Deng et~al.(2013)Deng, Hinton, and Kingsbury}]{deng2013new}
Li~Deng, Geoffrey Hinton, and Brian Kingsbury. 2013.
\newblock New types of deep neural network learning for speech recognition and
  related applications: An overview.
\newblock In \emph{2013 IEEE International Conference on Acoustics, Speech and
  Signal Processing}.

\bibitem[{Devlin et~al.(2019)Devlin, Chang, Lee, and
  Toutanova}]{devlin2019bert}
Jacob Devlin, Ming-Wei Chang, Kenton Lee, and Kristina Toutanova. 2019.
\newblock Bert: Pre-training of deep bidirectional transformers for language
  understanding.
\newblock In \emph{Proceedings of the 2019 Conference of the North American
  Chapter of the Association for Computational Linguistics: Human Language
  Technologies}, pages 4171--4186.

\bibitem[{Dong et~al.(2015)Dong, Wu, He, Yu, and Wang}]{dong2015multi}
Daxiang Dong, Hua Wu, Wei He, Dianhai Yu, and Haifeng Wang. 2015.
\newblock Multi-task learning for multiple language translation.
\newblock In \emph{Proceedings of the 53rd Annual Meeting of the Association
  for Computational Linguistics}, pages 1723--1732.

\bibitem[{Elman(1993)}]{elman1993learning}
Jeffrey~L Elman. 1993.
\newblock Learning and development in neural networks: The importance of
  starting small.
\newblock \emph{Cognition}, pages 71--99.

\bibitem[{Eriguchi et~al.(2017)Eriguchi, Tsuruoka, and
  Cho}]{eriguchi2017learning}
Akiko Eriguchi, Yoshimasa Tsuruoka, and Kyunghyun Cho. 2017.
\newblock Learning to parse and translate improves neural machine translation.
\newblock In \emph{Proceedings of the 55th Annual Meeting of the Association
  for Computational Linguistics}, pages 72--78.

\bibitem[{Firat et~al.(2016{\natexlab{a}})Firat, Cho, and
  Bengio}]{firat2016multi}
Orhan Firat, Kyunghyun Cho, and Yoshua Bengio. 2016{\natexlab{a}}.
\newblock Multi-way, multilingual neural machine translation with a shared
  attention mechanism.
\newblock In \emph{Proceedings of the 2016 Conference of the North American
  Chapter of the Association for Computational Linguistics: Human Language
  Technologies}, pages 866--875.

\bibitem[{Firat et~al.(2016{\natexlab{b}})Firat, Sankaran, Al-Onaizan, Vural,
  and Cho}]{firat2016zero}
Orhan Firat, Baskaran Sankaran, Yaser Al-Onaizan, Fatos T~Yarman Vural, and
  Kyunghyun Cho. 2016{\natexlab{b}}.
\newblock Zero-resource translation with multi-lingual neural machine
  translation.
\newblock In \emph{Proceedings of the 2016 Conference on Empirical Methods in
  Natural Language Processing}, pages 268--277.

\bibitem[{Gehring et~al.(2017)Gehring, Auli, Grangier, Yarats, and
  Dauphin}]{gehring2017convolutional}
Jonas Gehring, Michael Auli, David Grangier, Denis Yarats, and Yann~N Dauphin.
  2017.
\newblock Convolutional sequence to sequence learning.
\newblock In \emph{Proceedings of the 34th International Conference on Machine
  Learning}, pages 1243--1252. JMLR. org.

\bibitem[{Gu et~al.(2018)Gu, Hassan, Devlin, and Li}]{gu2018universal}
Jiatao Gu, Hany Hassan, Jacob Devlin, and Victor~OK Li. 2018.
\newblock Universal neural machine translation for extremely low resource
  languages.
\newblock In \emph{Proceedings of the 2018 Conference of the North American
  Chapter of the Association for Computational Linguistics: Human Language
  Technologies}, pages 344--354.

\bibitem[{Ha et~al.(2016)Ha, Niehues, and Waibel}]{ha2016toward}
Thanh-Le Ha, Jan Niehues, and Alexander Waibel. 2016.
\newblock Toward multilingual neural machine translation with universal encoder
  and decoder.
\newblock \emph{arXiv preprint arXiv:1611.04798}.

\bibitem[{He et~al.(2016)He, Xia, Qin, Wang, Yu, Liu, and Ma}]{he2016dual}
Di~He, Yingce Xia, Tao Qin, Liwei Wang, Nenghai Yu, Tie-Yan Liu, and Wei-Ying
  Ma. 2016.
\newblock Dual learning for machine translation.
\newblock In \emph{Advances in Neural Information Processing Systems}, pages
  820--828.

\bibitem[{Johnson et~al.(2017)Johnson, Schuster, Le, Krikun, Wu, Chen, Thorat,
  Vi{\'e}gas, Wattenberg, Corrado et~al.}]{johnson2017google}
Melvin Johnson, Mike Schuster, Quoc~V Le, Maxim Krikun, Yonghui Wu, Zhifeng
  Chen, Nikhil Thorat, Fernanda Vi{\'e}gas, Martin Wattenberg, Greg Corrado,
  et~al. 2017.
\newblock Google’s multilingual neural machine translation system: Enabling
  zero-shot translation.
\newblock \emph{Transactions of the Association for Computational Linguistics},
  pages 339--351.

\bibitem[{Kingma and Ba(2015)}]{kingma2014adam}
Diederik~P Kingma and Jimmy Ba. 2015.
\newblock Adam: A method for stochastic optimization.
\newblock \emph{International Conference on Learning Representations}.

\bibitem[{Kiperwasser and Ballesteros(2018)}]{kiperwasser2018scheduled}
Eliyahu Kiperwasser and Miguel Ballesteros. 2018.
\newblock Scheduled multi-task learning: From syntax to translation.
\newblock \emph{Transactions of the Association for Computational Linguistics},
  pages 225--240.

\bibitem[{Kudo and Richardson(2018)}]{kudo2018sentencepiece}
Taku Kudo and John Richardson. 2018.
\newblock Sentencepiece: A simple and language independent subword tokenizer
  and detokenizer for neural text processing.
\newblock \emph{arXiv preprint arXiv:1808.06226}.

\bibitem[{Lample et~al.(2018{\natexlab{a}})Lample, Conneau, Denoyer, and
  Ranzato}]{lample2018unsupervised}
Guillaume Lample, Alexis Conneau, Ludovic Denoyer, and Marc'Aurelio Ranzato.
  2018{\natexlab{a}}.
\newblock Unsupervised machine translation using monolingual corpora only.
\newblock In \emph{International Conference on Learning Representations}.

\bibitem[{Lample et~al.(2018{\natexlab{b}})Lample, Ott, Conneau, Denoyer, and
  Ranzato}]{lample2018phrase}
Guillaume Lample, Myle Ott, Alexis Conneau, Ludovic Denoyer, and Marc’Aurelio
  Ranzato. 2018{\natexlab{b}}.
\newblock Phrase-based \& neural unsupervised machine translation.
\newblock In \emph{Proceedings of the 2018 Conference on Empirical Methods in
  Natural Language Processing}, pages 5039--5049.

\bibitem[{Lewis et~al.(2019)Lewis, Liu, Goyal, Ghazvininejad, Mohamed, Levy,
  Stoyanov, and Zettlemoyer}]{lewis2019bart}
Mike Lewis, Yinhan Liu, Naman Goyal, Marjan Ghazvininejad, Abdelrahman Mohamed,
  Omer Levy, Ves Stoyanov, and Luke Zettlemoyer. 2019.
\newblock Bart: Denoising sequence-to-sequence pre-training for natural
  language generation, translation, and comprehension.
\newblock \emph{arXiv preprint arXiv:1910.13461}.

\bibitem[{Liang et~al.(2020)Liang, Duan, Gong, Wu, Guo, Qi, Gong, Shou, Jiang,
  Cao, Fan, Zhang, Agrawal, Cui, Wei, Bharti, Qiao, Chen, Wu, Liu, Yang,
  Campos, Majumder, and Zhou}]{XGLUE}
Yaobo Liang, Nan Duan, Yeyun Gong, Ning Wu, Fenfei Guo, Weizhen Qi, Ming Gong,
  Linjun Shou, Daxin Jiang, Guihong Cao, Xiaodong Fan, Ruofei Zhang, Rahul
  Agrawal, Edward Cui, Sining Wei, Taroon Bharti, Ying Qiao, Jiun-Hung Chen,
  Winnie Wu, Shuguang Liu, Fan Yang, Daniel Campos, Rangan Majumder, and Ming
  Zhou. 2020.
\newblock Xglue: A new benchmark dataset for cross-lingual pre-training,
  understanding and generation.
\newblock \emph{arXiv}, abs/2004.01401.

\bibitem[{Liu et~al.(2020)Liu, Gu, Goyal, Li, Edunov, Ghazvininejad, Lewis, and
  Zettlemoyer}]{liu2020multilingual}
Yinhan Liu, Jiatao Gu, Naman Goyal, Xian Li, Sergey Edunov, Marjan
  Ghazvininejad, Mike Lewis, and Luke Zettlemoyer. 2020.
\newblock Multilingual denoising pre-training for neural machine translation.
\newblock \emph{arXiv preprint arXiv:2001.08210}.

\bibitem[{Luong et~al.(2015)Luong, Le, Sutskever, Vinyals, and
  Kaiser}]{luong2015multi}
Minh-Thang Luong, Quoc~V Le, Ilya Sutskever, Oriol Vinyals, and Lukasz Kaiser.
  2015.
\newblock Multi-task sequence to sequence learning.
\newblock \emph{arXiv preprint arXiv:1511.06114}.

\bibitem[{Nguyen and Chiang(2017)}]{nguyen2017transfer}
Toan~Q Nguyen and David Chiang. 2017.
\newblock Transfer learning across low-resource, related languages for neural
  machine translation.
\newblock In \emph{Proceedings of the Eighth International Joint Conference on
  Natural Language Processing}, pages 296--301.

\bibitem[{Niehues and Cho(2017)}]{niehues2017exploiting}
Jan Niehues and Eunah Cho. 2017.
\newblock Exploiting linguistic resources for neural machine translation using
  multi-task learning.
\newblock In \emph{Proceedings of the Second Conference on Machine
  Translation}, pages 80--89.

\bibitem[{Ott et~al.(2019)Ott, Edunov, Baevski, Fan, Gross, Ng, Grangier, and
  Auli}]{ott2019fairseq}
Myle Ott, Sergey Edunov, Alexei Baevski, Angela Fan, Sam Gross, Nathan Ng,
  David Grangier, and Michael Auli. 2019.
\newblock fairseq: A fast, extensible toolkit for sequence modeling.
\newblock In \emph{Proceedings of NAACL-HLT 2019: Demonstrations}.

\bibitem[{Post(2018)}]{post2018call}
Matt Post. 2018.
\newblock A call for clarity in reporting bleu scores.
\newblock In \emph{Conference on Machine Translation}.

\bibitem[{Radford et~al.(2018)Radford, Narasimhan, Salimans, and
  Sutskever}]{radford2018improving}
Alec Radford, Karthik Narasimhan, Tim Salimans, and Ilya Sutskever. 2018.
\newblock Improving language understanding by generative pre-training.

\bibitem[{Ruder(2017)}]{ruder2017overview}
Sebastian Ruder. 2017.
\newblock An overview of multi-task learning in deep neural networks.
\newblock \emph{arXiv preprint arXiv:1706.05098}.

\bibitem[{Sen et~al.(2019)Sen, Gupta, Ekbal, and
  Bhattacharyya}]{sen2019multilingual}
Sukanta Sen, Kamal~Kumar Gupta, Asif Ekbal, and Pushpak Bhattacharyya. 2019.
\newblock Multilingual unsupervised nmt using shared encoder and
  language-specific decoders.
\newblock In \emph{Proceedings of the 57th Annual Meeting of the Association
  for Computational Linguistics}, pages 3083--3089.

\bibitem[{Sennrich et~al.(2016)Sennrich, Haddow, and
  Birch}]{sennrich2016improving}
Rico Sennrich, Barry Haddow, and Alexandra Birch. 2016.
\newblock Improving neural machine translation models with monolingual data.
\newblock In \emph{Proceedings of the 54th Annual Meeting of the Association
  for Computational Linguistics}, pages 86--96.

\bibitem[{Siddhant et~al.(2020)Siddhant, Johnson, Tsai, Ari, Riesa, Bapna,
  Firat, and Raman}]{Siddhant2020}
Aditya Siddhant, Melvin Johnson, Henry Tsai, Naveen Ari, Jason Riesa, Ankur
  Bapna, Orhan Firat, and Karthik Raman. 2020.
\newblock \href {https://aaai.org/ojs/index.php/AAAI/article/view/6414}
  {Evaluating the cross-lingual effectiveness of massively multilingual neural
  machine translation}.
\newblock In \emph{The Thirty-Fourth {AAAI} Conference on Artificial
  Intelligence, {AAAI} 2020, The Thirty-Second Innovative Applications of
  Artificial Intelligence Conference, {IAAI} 2020, The Tenth {AAAI} Symposium
  on Educational Advances in Artificial Intelligence, {EAAI} 2020, New York,
  NY, USA, February 7-12, 2020}, pages 8854--8861. {AAAI} Press.

\bibitem[{Sutskever et~al.(2014)Sutskever, Vinyals, and
  Le}]{sutskever2014sequence}
Ilya Sutskever, Oriol Vinyals, and Quoc~V Le. 2014.
\newblock Sequence to sequence learning with neural networks.
\newblock In \emph{Advances in neural information processing systems}, pages
  3104--3112.

\bibitem[{Szegedy et~al.(2016)Szegedy, Vanhoucke, Ioffe, Shlens, and
  Wojna}]{szegedy2016rethinking}
Christian Szegedy, Vincent Vanhoucke, Sergey Ioffe, Jon Shlens, and Zbigniew
  Wojna. 2016.
\newblock Rethinking the inception architecture for computer vision.
\newblock In \emph{Proceedings of the IEEE conference on computer vision and
  pattern recognition}, pages 2818--2826.

\bibitem[{Thompson et~al.(2019)Thompson, Gwinnup, Khayrallah, Duh, and
  Koehn}]{thompson2019overcoming}
Brian Thompson, Jeremy Gwinnup, Huda Khayrallah, Kevin Duh, and Philipp Koehn.
  2019.
\newblock Overcoming catastrophic forgetting during domain adaptation of neural
  machine translation.
\newblock In \emph{Proceedings of the 2019 Conference of the North American
  Chapter of the Association for Computational Linguistics: Human Language
  Technologies}, pages 2062--2068.

\bibitem[{Utiyama and Isahara(2007)}]{utiyama2007comparison}
Masao Utiyama and Hitoshi Isahara. 2007.
\newblock A comparison of pivot methods for phrase-based statistical machine
  translation.
\newblock In \emph{Human Language Technologies 2007: The Conference of the
  North American Chapter of the Association for Computational Linguistics;
  Proceedings of the Main Conference}, pages 484--491.

\bibitem[{Vaswani et~al.(2017)Vaswani, Shazeer, Parmar, Uszkoreit, Jones,
  Gomez, Kaiser, and Polosukhin}]{vaswani2017attention}
Ashish Vaswani, Noam Shazeer, Niki Parmar, Jakob Uszkoreit, Llion Jones,
  Aidan~N Gomez, {\L}ukasz Kaiser, and Illia Polosukhin. 2017.
\newblock Attention is all you need.
\newblock In \emph{Advances in neural information processing systems}.

\bibitem[{Vincent et~al.(2008)Vincent, Larochelle, Bengio, and
  Manzagol}]{vincent2008extracting}
Pascal Vincent, Hugo Larochelle, Yoshua Bengio, and Pierre-Antoine Manzagol.
  2008.
\newblock Extracting and composing robust features with denoising autoencoders.
\newblock In \emph{Proceedings of the 25th international conference on Machine
  learning}, pages 1096--1103.

\bibitem[{Wang et~al.(2019)Wang, Xia, He, Tian, Qin, Zhai, and
  Liu}]{wang2018multiagent}
Yiren Wang, Yingce Xia, Tianyu He, Fei Tian, Tao Qin, ChengXiang Zhai, and
  Tie-Yan Liu. 2019.
\newblock Multi-agent dual learning.
\newblock In \emph{International Conference on Learning Representations}.

\bibitem[{Wenzek et~al.(2019)Wenzek, Lachaux, Conneau, Chaudhary, Guzman,
  Joulin, and Grave}]{wenzek2019ccnet}
Guillaume Wenzek, Marie-Anne Lachaux, Alexis Conneau, Vishrav Chaudhary,
  Francisco Guzman, Armand Joulin, and Edouard Grave. 2019.
\newblock Ccnet: Extracting high quality monolingual datasets from web crawl
  data.
\newblock \emph{arXiv preprint arXiv:1911.00359}.

\bibitem[{Wu et~al.(2019)Wu, Wang, Xia, Tao, Lai, and Liu}]{wu2019exploiting}
Lijun Wu, Yiren Wang, Yingce Xia, QIN Tao, Jianhuang Lai, and Tie-Yan Liu.
  2019.
\newblock Exploiting monolingual data at scale for neural machine translation.
\newblock In \emph{Proceedings of the 2019 Conference on Empirical Methods in
  Natural Language Processing and the 9th International Joint Conference on
  Natural Language Processing (EMNLP-IJCNLP)}, pages 4198--4207.

\bibitem[{Zaremoodi and Haffari(2018)}]{zaremoodi2018neural}
Poorya Zaremoodi and Gholamreza Haffari. 2018.
\newblock Neural machine translation for bilingually scarce scenarios: a deep
  multi-task learning approach.
\newblock In \emph{Proceedings of the 2018 Conference of the North American
  Chapter of the Association for Computational Linguistics: Human Language
  Technologies}.

\bibitem[{Zhang and Zong(2016)}]{zhang2016exploiting}
Jiajun Zhang and Chengqing Zong. 2016.
\newblock Exploiting source-side monolingual data in neural machine
  translation.
\newblock In \emph{Proceedings of the 2016 Conference on Empirical Methods in
  Natural Language Processing}, pages 1535--1545.

\bibitem[{Zoph et~al.(2016)Zoph, Yuret, May, and Knight}]{zoph2016transfer}
Barret Zoph, Deniz Yuret, Jonathan May, and Kevin Knight. 2016.
\newblock Transfer learning for low-resource neural machine translation.
\newblock In \emph{Proceedings of the 2016 Conference on Empirical Methods in
  Natural Language Processing}, pages 1568--1575.

\end{thebibliography}

\appendix

\clearpage
\section*{Appendices}

\begin{subappendices}
\section{Bitext Training Data}
\label{appen:bitext}
We concatenate all resources except WikiTitles provided by WMT of the latest available year and filter out duplicated pairs and pairs with the same source and target sentence. For Fr and Cs, we randomly sample $10$M sentence pairs from the full corpus. The detailed statistics of bitext data can be found in Table~\ref{tbl:data-stats}. 

We randomly sample $1,000$ sentence pairs from each individual validation set and concatenate them to construct a multilingual validation set. We tokenize all data with the SentencePiece model~\cite{kudo2018sentencepiece}, forming a vocabulary shared by all the source and target languages with $32$k tokens for bilingual models ($16$k for Hi and Gu) and $64$k tokens for multilingual models. 

\begin{table}[h]
\centering
\begin{tabular}{cccc}
\toprule
Code & Language & \#Bitext   & Validation   \\
\midrule
Fr   & French   & $10$M      & Newstest$13$ \\
Cs   & Czech    & $10$M      & Newstest$16$ \\
De   & German   & $4.6$M     & Newstest$16$ \\
Fi   & Finnish  & $4.8$M     & Newstest$16$ \\
Lv   & Latvian  & $1.4$M     & Newsdev$17$  \\
Et   & Estonian & $0.7$M     & Newsdev$18$  \\
Ro   & Romanian & $0.5$M     & Newsdev$16$  \\
Hi   & Hindi    & $0.26$M    & Newsdev$14$  \\
Tr   & Turkish  & $0.18$M    & Newstest$16$ \\
Gu   & Gujarati & $0.08$M    & Newsdev$19$  \\
\bottomrule
\end{tabular}
\caption{Statistics of the parallel resources from WMT. A list of 10 languages ranked with the size of bitext corpus translating to/from English. \label{tbl:data-stats}}
\end{table}
\end{subappendices}
\end{document}